\documentclass[fp,twocolumn]{jpsj3}
\usepackage{txfonts}
\usepackage{multirow}
\usepackage{here}
\usepackage{color}

\title{Semi-supervised learning combining backpropagation and STDP: STDP enhances learning by backpropagation with a small amount of labeled data in a spiking neural network}

\author{Kotaro Furuya$^1$  and Jun Ohkubo$^2$$^,$$^3$}
\inst{$^1$School of Computing, Tokyo Institute of Technology, 2-12-1 Ookayama, Meguro-ku, Tokyo 152-8552, Japan \\
$^2$Graduate School of Science and Engineering, Saitama University, 255 Shimo-Okubo, Sakura-ku, Saitama-shi
338-8570, Japan\\
$^3$JST, PREST, 4-1-8 Honcho, Kawaguchi, Saitama 332-0012, Japan} 

\abst{A semi-supervised learning method for spiking neural networks is proposed. The proposed method consists of supervised learning by backpropagation and subsequent unsupervised learning by spike-timing-dependent plasticity (STDP), which is a biologically plausible learning rule. Numerical experiments show that the proposed method improves the accuracy without additional labeling when a small amount of labeled data is used. This feature has not been achieved by existing semi-supervised learning methods of discriminative models. It is possible to implement the proposed learning method for event-driven systems. Hence, it would be highly efficient in real-time problems if it were implemented on neuromorphic hardware. The results suggest that STDP plays an important role other than self-organization when applied after supervised learning, which differs from the previous method of using STDP as pre-training interpreted as self-organization.}

\begin{document}
\maketitle

\section{Introduction}
The brain is a large network of neurons. The neurons are connected by synapses, and information is transmitted by spikes. A spike is a short pulse signal, and its waveform does not carry any information. Rather, the number and timing of spikes are thought to transmit information.\cite{NeuronalDynamics}
In recent years, artificial neural networks (ANNs) inspired by this structure of the brain have achieved great success when applied to various problems of machine learning using techniques such as deep learning.\cite{lecun2015deep}
However, the transmission of information in ANNs is analog-valued, such as outputs of the sigmoid function, and ANNs do not transmit spikes like real neurons in the brain. 
Furthermore, ANNs do not have the dynamics that the membrane potential rises in time series until it exceeds the threshold to fire. Therefore, they are very different from real brains.
As an alternative, spiking neural networks (SNNs), which offer high energy efficiency and performance in real-time problems when used with neuromorphic hardware,\cite{Efficient_BP_Neuromophic} have been proposed.\cite{maass1997SNN}
SNNs use models that focus on action potentials, such as the integrate-and-fire model, to transmit information by increasing the membrane potential and firing in response to each input at a specific time.
It is known that the output spike trains are sparse in time and each spike has high information content.\cite{Review_deepL_inSNN}
Thus, SNNs are more biologically plausible, capable of detailed behaviors, and energy efficient.
In recent years, research that combines fields such as high-energy physics with machine learning, especially deep learning in ANNs, has been actively conducted.\cite{Guest_2018}
Even in fields of physics in which ANNs have shown promising results, SNNs are expected to enable the application of advanced and accurate algorithms owing to their high efficiency in real-time processing.\cite{Borzyszkowski:2687102}

Despite theoretically demonstrating that the computational power of SNNs is at least equal to that of ANNs,\cite{maass2004computational} their performance has been significantly inferior to that of ANNs in various machine learning problems.
The reason for this poor performance is that there has been no suitable learning method for SNNs.
The spike is a discontinuous process and is nondifferentiable. 
Therefore, we cannot use straightforwardly error backpropagation (BP), a very powerful supervised learning algorithm that is widely used in ANNs.
However, in recent years, many approximate methods of BP have been proposed for SNNs, and they are beginning to show performance close to that of ANNs.\cite{SpikeProp,TrainingDeepSNN2020,TemporalBackprop,Lee_SpikingNNBackprop}
For example, Bohte et al. treated the firing time as a nonlinear function based on the membrane potential,\cite{SpikeProp} Ledinauskas et al. tuned a surrogate gradient function,\cite{TrainingDeepSNN2020} Mostafa used a transformation of time variables,\cite{TemporalBackprop} and Lee et al. treated nondifferentiable points as noise.\cite{Lee_SpikingNNBackprop} In each of these studies, SNNs were trained by BP, and high performance was achieved.

In addition, because neurons of SNNs fire in time series and transmit information using spikes, they can be trained by spike-timing-dependent plasticity (STDP), which is a biologically plausible unsupervised learning mechanism.
STDP is a phenomenon discovered by observing biological synapses that manipulates the strength of synaptic connections between connected neurons depending on pre- and post-synaptic spike timings.\cite{STDP_Overview}
Several studies have reported that unsupervised learning, such as image classification, can be performed by applying STDP to SNNs. \cite{iakymchuk2015simplified,STDP_MNIST}
STDP is an unsupervised learning rule, but it can be used for supervised learning with various mechanisms.\cite{supervisedSTDP,BP-STDP}

In supervised learning, such as learning by BP, a large amount of labeled data is required to obtain high performance.
However, it is not easy to obtain labeled data because manual labeling is expensive.
In contrast, in unsupervised learning such as learning by STDP, labels are not required. Unlabeled data are easy to obtain. Hence, in the field of machine learning, semi-supervised learning that uses both labeled and unlabeled data has been developed to reduce the cost of data labeling.\cite{semi-supervisedReview}
In semi-supervised learning, a small amount of labeled data and a large amount of unlabeled data are generally used for learning.
Although self-training is a method of discriminative models in semi-supervised learning, the accuracy is improved by labeling unlabeled data and increasing the amount of labeled data.
In such a methodology, if the labeling is incorrect, the learning will fail.\cite{survey_suemi-supervised}

In ANNs, layer-wise unsupervised pre-training followed by supervised fine-tuning can improve the accuracy, and this approach has achieved high performance.\cite{erhan2010does}
Such a pre-training methodology was introduced by Hinton et al.\cite{hinton2006fast}. 
In that study, the deep brief network was trained by greedy layer-wise training.
In the spiking domain, some methods have been proposed in which STDP is applied as unsupervised pre-training, followed by fine-tuning using BP.\cite{SNN_STDP_followedby_BPFineTunig,Semi-Supervised_BP_after_STDP}
Lee et al. succeeded in improving the robustness and speeding up the learning procedure using pre-training by STDP.\cite{SNN_STDP_followedby_BPFineTunig} Dorogyy and Kolisnichenko reported that the accuracy is improved by pre-training with STDP when a small amount of labeled data is used.\cite{Semi-Supervised_BP_after_STDP}
The flow of supervised learning after STDP-based pre-training is thought to model the situation in which living things observe the world around them and are then taught about a certain thing such as a name by their teachers.
In contrast, we can also consider the situation in which after being taught a certain thing, they observe it many times to improve their understanding. In other words, the sequence of STDP-based unsupervised learning following supervised learning also models real-world situations, and such research has not been previously reported.

In this paper, we propose a semi-supervised learning method consisting of STDP-based unsupervised learning after BP-based supervised learning.
Numerical experiments to solve the task of handwritten digit recognition were performed to evaluate the capability of the proposed method.
The results show that the proposed method improves accuracy when a small amount of labeled data is available and solves the problem of self-training because there is no need to label unlabeled data.

The remainder of this paper is organized as follows.
In Sect. 2, we present the prior knowledge necessary to explain and provide the context for the proposed method.
Section 3 describes our proposed method, Sect. 4 presents the numerical results, Sect. 5 discusses the results, and Sect. 6 summarizes the conclusions of this study.

\section{Prior Knowledge}

\subsection{Spiking neural network}
In this study, we use fully connected feed-forward SNNs, in which each neuron is connected to all neurons in the next layer and information transfers from the input layer to the output layer in one direction.
In addition, we assume that each layer has a winner-take-all (WTA) circuit, which is described in Sect. 2.1.2.\cite{oster2009WinnerTakeALL}
Let $N$ be the number of neurons and $M$ be the number of synapses in a layer of the SNN. The number of active neurons that output spikes is denoted by $n$, and the number of active synapses that receive input spikes is denoted by $m$.
In addition, a variable $x$ in the $l$-th layer is expressed as $x^{(l)}$.

\subsubsection{Leaky integrate-and-fire neuron}
The leaky integrate-and-fire (LIF) neuron, one of the best-known spiking neural models\cite{gerstner2002SpikingNeuronModel}, integrates the input spikes weighted by the strength of synaptic connections and changes its membrane potential accordingly.
This model approximates the biological phenomenon in which the membrane potential rises with time integration of the input, and when the membrane potential is above a critical voltage, an action potential is triggered and the membrane potential is reset. 
Because the state of the LIF model is updated only when an input is received, the update of the membrane potential for a given input spike is expressed as follows:\cite{Lee_SpikingNNBackprop}
\begin{equation}
	V_{\rm mp}(t_p)=V_{\rm mp}(t_{p-1})e^{\frac{t_{p-1}-t_p}{\tau_{{\rm mp}}}}+w_i w_{{\rm dyn}},
    \label{eq:membrane update}
\end{equation}
where $V_{\rm mp}$ is the membrane potential, $\tau_{\rm mp}$ is the membrane time constant, $t_p$ and $t_{p-1}$ are the $p$ and $p$$-$$1$-th input times (that is, the present and previous input times), and $ w_i$ is the weight of the $i$-th synapse that the input spike passes through. $w_{\rm dyn}$ is a variable that controls the refractory period and is expressed as follows: 
\begin{equation}
	w_{\rm dyn}=\left\{ 
    \begin{array}{ll}
    \left(\frac{\Delta_t}{T_{\rm ref}}\right)^2 & {\rm if}\ \Delta_t<T_{\rm ref},\\
    1 & \rm otherwise,
    \end{array}
    \right. \nonumber
\end{equation}
where $T_{\rm ref}$ is the maximum refractory period. 
We define $\Delta_t=t_{\rm out}-t_p$, where $t_{\rm out}$ is the most recent firing time of the neuron.
The refractory period is a period during which a neuron does not respond to the stimulus (input) immediately after spiking. 
When the membrane potential $V_{\rm mp}$ exceeds the threshold $V_{\rm th}$, the LIF neuron generates a spike and the membrane potential decreases sharply:
\begin{equation}
	 V_{\rm mp}(t_p^+)=V_{\rm mp}(t_p)-V_{\rm th},
     \label{eq:reset}
\end{equation}
where $t_p^+$ is the time immediately after spiking.

\subsubsection{Winner-take-all circuit}
\begin{figure}
    \begin{center}
         \includegraphics[scale=0.4]{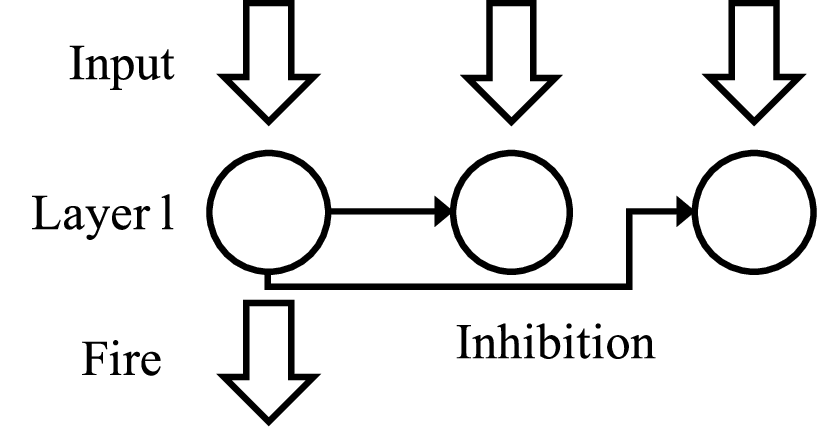}
        \caption{Conceptual diagram of a winner-take-all circuit.}
        \label{fig:WTA}
    \end{center}
\end{figure}

A WTA circuit is a principle generally used in recurrent neural networks that inhibits the firing of other neurons in the same layer when one neuron fires.\cite{oster2009WinnerTakeALL}
A conceptual diagram of a WTA circuit is shown in Fig. \ref{fig:WTA}.
Biologically, such circuits play a role in cortical processing models such as a hierarchical model of vision in the cortex.\cite{riesenhuber1999hierarchical}
In SNNs, it is used to improve accuracy, stability, and learning speed as well as in unsupervised learning with STDP.

\subsection{Visual receptive field}
Organisms perceive changes in themselves and their environment by stimulus, process the stimuli, and communicate with others.
However, they can only respond to stimuli in a specific zone.
This limited area is called the receptive field.\cite{NeuronalDynamics}
The receptive field of a visual neuron refers to the area of the retina where cells can react to light.
There are various types of receptive fields in the visual system, for example, on-center and off-center receptive fields.
An on-center receptive field shows an excitatory response when stimulated at the center of the receptive field and an inhibitory response when stimulated at the peripheral part. 
In contrast, the off-center receptive field shows an inhibitory response when stimulated centrally and an excitatory response when stimulated peripherally.

The elements of the response matrix $R$ of a receptive field can be expressed as the Frobenius inner product of the submatrix of the input stimulus matrix $S$ and the receptive field structure matrix $F$, and are formulated as follows: 
\begin{equation}
	R_{ij} = \sum_{k_1=0}^{K-1} \sum_{k_2=0}^{K-1} S_{(i+{k_1})(j+{k_2})}F_{{k_1}{k_2}},
\end{equation}
where $S \in \mathbb{R}^{H \times W}, F \in \mathbb{R}^{K \times K},$ and $R \in  \mathbb{R}^{(H-K+1) \times (W-K+1)}$.
$H$ and $W$ are the height and width of the input stimulus matrix $S$, respectively, and $K$ is the size of the receptive field structure matrix $F$.

\subsection{Backpropagation in SNNs}
\label{subsec:backprop}
The spikes that carry information in SNNs are not differentiable because of their discontinuity.
Therefore, to calculate the derivative required for the BP algorithm, an approximation or other method must be used.
In this section, we introduce an approximation method that enables the derivation of differentiable transfer functions and their derivatives in SNNs.\cite{Lee_SpikingNNBackprop}

First, we define two variables: $x_k(t)$, which is the accumulated effect of the $k$-th active input synapse onto the membrane potential of a target neuron, and $a_i(t)$, which is the generation of spikes in neuron $i$ acting on its own membrane potential.
These two variables are defined as sums of terms with exponential decays:
\begin{eqnarray}
    x_k(t) & = & \sum_p \mathrm{exp}\left(\frac{t_p-t}{\tau_{\rm mp}}\right), \label{eq:accumlated_x} \\ 
    a_i(t) & = & \sum_q \mathrm{exp}\left(\frac{t_q-t}{\tau_{\rm mp}}\right). \label{eq:accumlated_a}
\end{eqnarray}
Note that these two summations have different meanings: the first sum is over all input spike times $t_p < t$ at the $k$-th input synapse, and the second sum is over the output spike times $t_q < t$ for $a_i$.
Using these definitions, in the method by Lee et al.\cite{Lee_SpikingNNBackprop}, ignoring the effect of refractory periods, the membrane potential of the $i$-th LIF neuron at time $t$ is expressed as follows because of the properties of LIF neurons and WTA circuits: 
\begin{equation}
    V_{\mathrm{mp},i}(t)=\sum_{k=1}^m w_{ik}x_k(t)-V_{\mathrm{th},i}a_i(t)+\sigma V_{\mathrm{th},i}\sum_{j=1,j\neq i}^n \kappa_{ij}a_j(t), 
    \label{eq:accumlated potential}
\end{equation}
where $w_{ik}$ is the weight of the synapse between the $k$-th neuron in the previous layer and the $i$-th neuron in the current layer.
In Eq.~\eqref{eq:accumlated potential}, the first term represents inputs, the second term represents membrane potential resets, and the third term represents lateral inhibitions by the WTA circuit.
$\kappa_{ij}$ is the strength of the lateral inhibition parameter by a WTA mechanism from neuron $j$ to neuron $i$ and in $[-1,0]$.
$\sigma$ is a parameter that controls the effect of lateral inhibition.
From Eq.~\eqref{eq:accumlated potential}, if all layers have the same time constant $\tau _{\rm mp}$, the output ($a_i$) from the current layer becomes the input ($x_i$) of the next layer. 
This is the basis for deriving an error BP algorithm via the chain rule.

Because there are discontinuous jumps, Eqs.~\eqref{eq:accumlated_x}, \eqref{eq:accumlated_a}, and \eqref{eq:accumlated potential} are not differentiable at these points.
However, in the method proposed by Lee et al.\cite{Lee_SpikingNNBackprop}, these nondifferentiable points are considered as noise, and these equations are treated as differentiable continuous signals.
Owing to this approximation, the calculation of gradients includes errors, but their results show that the influence of these errors is very small.

The transfer function of the LIF neuron in WTA circuits is approximated as follows by setting the residual $V_{\mathrm{mp}}$ term to zero: 
\begin{equation}
  a_i \approx \frac{s_i}{V_{\mathrm{th},i}} + \sigma \sum _{j=1,j\neq i}^n \kappa_{ij}a_j,\label{eq:kasseika} 
\end{equation}
where
\begin{equation}
  s_i = \sum_{k=1}^m w_{ik}x_k . \nonumber
\end{equation}
We derive the following equations by directly differentiating Eq.~\eqref{eq:kasseika}:
\begin{eqnarray}
	\frac{\partial a_i}{\partial s_i} & \approx &  \frac{1}{V_{{\rm th},i}}, \ \  \frac{\partial a_i}{\partial w_{ik}} \approx \frac{\partial a_i}{\partial s_i} x_k, \nonumber \\
    \frac{\partial a_i}{\partial V_{{\rm th},i}} & \approx & \frac{\partial a_i}{\partial s_i} (-a_i+\sigma \sum _{j,j\neq i}^n \kappa_{ij}a_j), \label{eq:threshold} \\
    \frac{\partial a_i}{\partial \kappa _{ih}} & \approx & \frac{\partial a_i}{\partial s_i}(\sigma V_{{\rm th},i}a_h), \nonumber \\
    \begin{bmatrix}
    \frac{\partial a_i}{\partial x_k} \\
    \vdots \\
    \frac{\partial a_n}{\partial x_k}
    \end{bmatrix}
    & \approx & \frac{1}{\sigma}
    \begin{bmatrix}
    q & \cdots & -\kappa _{1n} \\
    \vdots & \ddots & \vdots \\
    -\kappa _{n1} & \cdots & q
    \end{bmatrix} ^ {-1}
    \begin{bmatrix}
    \frac{w_{1k}}{V_{{\rm th},1}} \\
    \vdots \\
    \frac{w_{nk}}{V_{{\rm th},n}}
    \label{eq:diff_1}
    \end{bmatrix} ,
\end{eqnarray}
where $q = \frac{1}{\sigma}$.
Here, assuming that the strengths of lateral inhibitions are all the same as $\mu$ ($\kappa _{ij}=\mu, \forall i,j$), we can simplify Eq.~\eqref{eq:diff_1} to
\begin{equation}
    \frac{\partial a_i}{\partial x_k}  \approx  \frac{\partial a_i}{\partial s_i} 
    \frac{1}{1-\mu \sigma}\left(w_{ik}-\frac{\mu \sigma V_{{\rm th},i}}{1+\mu \sigma(n-1)}\sum _{j=1}^n \frac{w_{jk}}{V_{{\rm th},j}}\right).
    \label{eq:diff}
\end{equation}
By substituting Eqs.~\eqref{eq:threshold} and \eqref{eq:diff} into the ordinary BP algorithm, it is possible to perform the BP algorithm in SNNs.
Further techniques such as initialization, normalization, and regularization have been performed in prior research to improve accuracy and efficiency.\cite{Lee_SpikingNNBackprop}

\subsection{Weight initialization and BP error normalization}
To deal with chaotic convergence behavior and facilitate stable training convergence in SNNs,  appropriate network initialization and optimization tools are important.\cite{SNN_STDP_followedby_BPFineTunig}
In this section, we describe the weight initialization and BP error normalization used in the error BP method introduced in Sect. \ref{subsec:backprop}.\cite{Lee_SpikingNNBackprop}

The thresholds of neurons and the weights of synaptic connections in the $l$-th layer are initialized as follows: 
\begin{equation}
    w^{(l)} \sim U\left[-\sqrt{3/M^{(l)}},\sqrt{3/M^{(l)}}\right], \ V_{\rm th}^{(l)} = \alpha \sqrt{3/M^{(l)}}, 
    \label{eq:weight_init}
\end{equation}
where $\alpha> 1$ is a constant. 
$U[-a,a]$ denotes a random number generated from a normal distribution in $(-a,a)$. The weight initialized by Eq.~\eqref{eq:weight_init} satisfies the following condition: 
\begin{equation}
    E\left[\sum_i ^{M^{(l)}} (w_{ji}^{(l)})^2 \right] = 1 \ \textrm{or} \ E\left[(w_{ji}^{(l)})^2 \right] = \frac{1}{M^{(l)}}.
    \label{eq:weight_condition}
\end{equation}
This condition is used in BP error normalization.

The main purpose of BP error normalization is to adjust the update the magnitudes of the weights and thresholds. 
In the  $l$-th layer, the error backpropagating through the $i$-th neuron is defined as follows: 
\begin{equation}
    \delta_i ^{(l)} = \frac{g_i^{(l)}}{\overline{g}^{(l)}} \sqrt{\frac{M^{(l+1)}}{m^{(l+1)}}} \sum _j ^{m^{(l+1)}} w_{ji}^{l+1}\delta_j ^{l+1},
\end{equation}
where $g_i^{(l)}=1/V_{{\rm th},i}^{(l)},\ \overline{g}^{(l)} =E \sqrt{\left[ (g_i^{(l)})^2 \right]} \simeq \sqrt{\frac{1}{n^{(l)}}\sum_i^{n^{(l)}}(g_i^{(l)})^2}$.
From Eq.~\eqref{eq:weight_condition}, the expected value of the squared sum of errors is constant for all layers and thus the updating of the weights and thresholds can be balanced.
Therefore, the weight and threshold are updated as
\begin{equation}
    \Delta w_{ij}^{(l)} = - \eta_{\rm w} \sqrt{\frac{M^{(l+1)}}{m^{(l)}}} \delta _i ^{(l)} x_j^{(l)},\ \Delta V_{{\rm th},i} = - \eta_{\rm th} \sqrt{\frac{M^{(l+1)}}{m^{(l)}M^{(l+1)}}} \delta _i ^{(l)} \hat{a}_i^{(l)},
\end{equation}
where $\eta_{\rm w}$ and $\eta_{\rm th}$ are the learning rates of the weight and threshold, respectively.
Let $\hat{a}_i = \gamma a_i-\sigma \sum_{j \neq i}^n \kappa_{ij} a_j$ and $\gamma$ be a parameter.
By performing the above normalization, in the initial stage of learning, the magnitude of the updates of weights and thresholds is determined according to the expected value of each active synapse, regardless of the number of active synapses and neurons.
Therefore, the update of all layers in SNNs can be balanced.

\subsection{Threshold regularization}
Threshold regularization is applied to improve the firing balance of neurons in SNNs.
This regularization has the effect of suppressing the generation of dead neurons and can improve the accuracy.
In particular, when the network has WTA mechanisms, lateral inhibition occurs at each layer, and thus threshold regularization is important. The details of the threshold regularization are described in this section.\cite{Lee_SpikingNNBackprop}

When $N_w$ neurons in a layer fire after receiving an input, the thresholds of the firing neurons are increased by $\rho N$, and the thresholds of all neurons in that layer are reduced by $\rho N_w$.
This process makes highly active neurons less sensitive to inputs because their thresholds increase, while less active neurons become more sensitive to inputs because their thresholds decrease. 
Therefore, the firing of neurons can be balanced, and the accuracy can be improved.

\begin{figure}[tb]
  \begin{center}
  \includegraphics[scale=0.3]{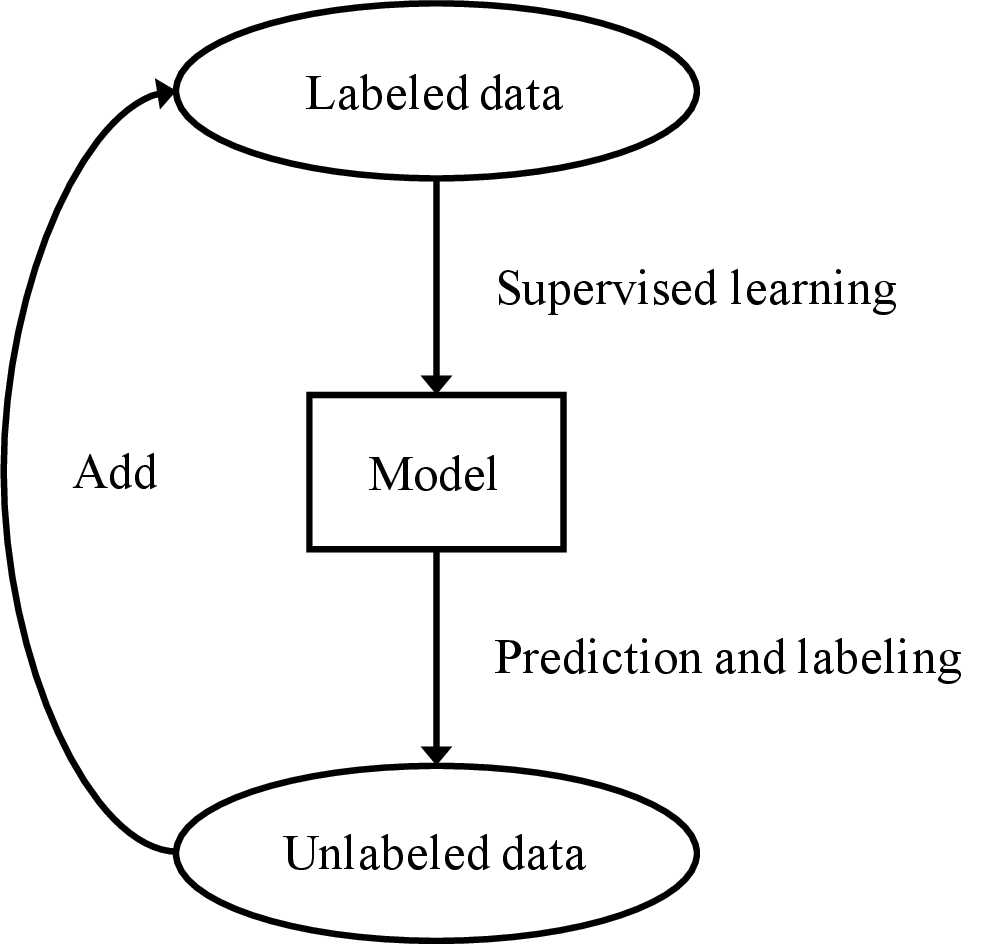}
  \caption{The flow of self-training.}
  \label{fig:ssl flow}
  \end{center}
\end{figure}

\subsection{Spike-timing-dependent plasticity}
\label{subsec:stdp}
STDP is a learning rule discovered by observing biological synapses that changes the strength of synaptic connections according to the temporal correlations of spikes between the connected neurons.\cite{STDP_Overview}
The outline of STDP learning is described in this section.\cite{STDP_formulation}
Although there are symmetric and asymmetric STDP rules for changing the strength of the synapse, we consider only an asymmetric one in this paper.

Here, we assume that neurons A and B are connected by a synapse and a spike signal is transmitted from neuron A to neuron B.
If neuron A fires and then neuron B fires within a certain time window, long-term potentiation (LTP) is triggered, and thus the synaptic connection between them is strengthened.
If the order is reversed, long-term depression (LTD) is triggered, and thus the synaptic connection between them is decreased.
We define the firing time of neuron A as $t_{\rm pre}$ and the firing time of neuron B as $t_{\rm post}$. Then, the timing difference between the pre- and post-synaptic spikes can be  expressed as $\Delta s =t_{\rm pre} - t_{\rm post}$.
The amount of synaptic modification $\Delta w$ can be written as
\begin{equation}
    \Delta w =\left\{
       \begin{array}{ll}
       A^+ {\rm exp}\left(\frac{\Delta s}{\tau _ {\rm plus}}\right)  & (\Delta s < 0), \\
       A^- {\rm exp}\left(\frac{- \Delta s}{\tau _ {\rm minus}}\right)  & (\Delta s \geq 0),
       \end{array}
       \right. 
\end{equation}
where $A^+$ and $A^-$ are positive and negative constants, respectively, which determine the maximum amount of synaptic change, and $\tau _{\rm plus}$ and $\tau _{\rm minus}$ are time constants.
The weight change is described with the STDP learning rate $\sigma _{\rm stdp}$:
\begin{equation}
    w_{\rm new} = w_{\rm old} + \sigma _{\rm stdp}\Delta w. 
    \label{eq:weight_change_stdp}
\end{equation}
We can train SNNs by applying the above unsupervised learning rule with WTA mechanisms.
In other words, it is possible to use the temporal correlations of neuronal firing for learning.

\section{Proposed Semi-supervised Learning Method}
\begin{figure}[tb]
    \begin{center}
    \includegraphics[scale=0.3]{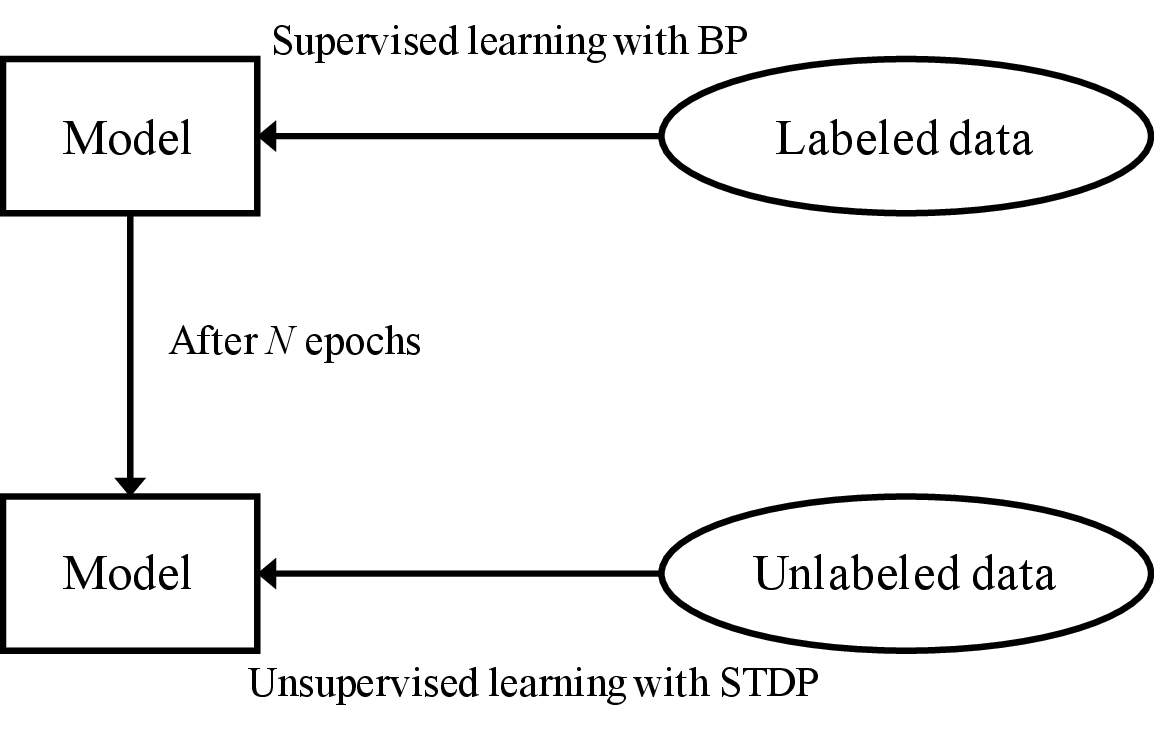}
    \caption{The flow of the proposed semi-supervised learning method.}
    \label{fig:proposed ssl flow}
    \end{center}
\end{figure}

In SNNs, some methods of using unsupervised learning by STDP as pre-training have been proposed,\cite{Semi-Supervised_BP_after_STDP,SNN_STDP_followedby_BPFineTunig} and the performance has been improved the same as pre-training in ANNs.
Such a learning flow can correspond to the situation in which organisms observe the world in advance and are then taught specific labels.
In contrast, we can assume a situation in which the labels are taught in advance and then subsequent observation deepens understanding.
Moreover, in the semi-supervised learning method based on the discriminative models, the accuracy is generally improved by labeling unlabeled data and re-learning. 
This methodology has a problem in that learning does not work properly if incorrect labeling is applied.\cite{survey_suemi-supervised}
The proposed method is a semi-supervised learning method in which unsupervised learning is performed by STDP after supervised learning by BP.
Because the proposed method improves accuracy by STDP, we do not need to label additional data. 
Also, labeling and re-learning are not repeated, and learning is completed in only two steps: learning by BP and learning by STDP.
Figures \ref{fig:ssl flow} and \ref{fig:proposed ssl flow} show the learning flows of self-training and the proposed method.
In this section, we describe the learning scheme of the proposed method and the network architecture used in the numerical experiments.

\subsection{Learning scheme}
In STDP, the temporal correlations of firing are used for learning. 
Thus, after learning to some extent by BP, learning by STDP can improve the results of learning using labeled data with learning using unlabeled data. This can model organisms observing their surroundings many times to improve their understanding after being taught something.
In our method, we first train the SNN by supervised learning by BP, and then unsupervised learning by STDP is applied.

The method of Lee et al.\cite{Lee_SpikingNNBackprop} described in Sect. \ref{subsec:backprop} was used as the BP method.
In addition, weight and threshold initialization and BP error normalization were performed according to Lee et al.\cite{Lee_SpikingNNBackprop}
Because the STDP-based unsupervised initialization scheme has an equivalent effect on learning as classic regularization techniques such as early stopping, L1/L2 weight decay, and dropout,\cite{SNN_STDP_followedby_BPFineTunig} we can expect a similar effect of STDP-based post learning.
Thus, weight regularization was not performed in our method to clarify the results.

As STDP learning rules, we use the following modification of the learning rules described in Sect. \ref{subsec:stdp}:
 \begin{equation}
 \Delta w =\left\{
    \begin{array}{ll}
    A^+ \rm exp\left(\frac{\Delta s}{\tau _{plus}}\right)  & (\Delta s \  \leq -1), \\
    A^- \rm exp\left(\frac{- \Delta s}{\tau _{minus}}\right)  & (\Delta s \geq 1). 
    \end{array}
    \right.
\end{equation}
Considering the synaptic conduction velocity, no update occurs when the time interval is less than 1 ms.
By making the above modifications, we can expect that the symmetry of updates will improve the accuracy because the weights are less biased in either the positive and negative directions and the firings of neurons are balanced.
In particular, when simulating in discrete time, as in our case, our numerical experiments showed that when $|{\Delta} s|$ is less than 1 ms, only positive updates occur, and thus the weights are biased and the accuracy does not improve compared to the case where the updates are symmetric.
We use a weight update formula similar to Eq.~\eqref{eq:weight_change_stdp}.
The time window is [1 ms, ..., 20 ms] in both positive and negative directions.
Based on the result of biological experiments,\cite{STDP_TimeWindow} the modification is assumed only when the interval is equal to 20 ms or less in both directions.
Threshold regularization is applied during forward propagation in BP and STDP learning. 
Threshold regularization is almost the same as that used in Lee et al.\cite{Lee_SpikingNNBackprop}, but the threshold is reduced only for the neurons that do not fire.

\begin{figure}[t]
    \begin{center}
    \includegraphics[scale=0.27]{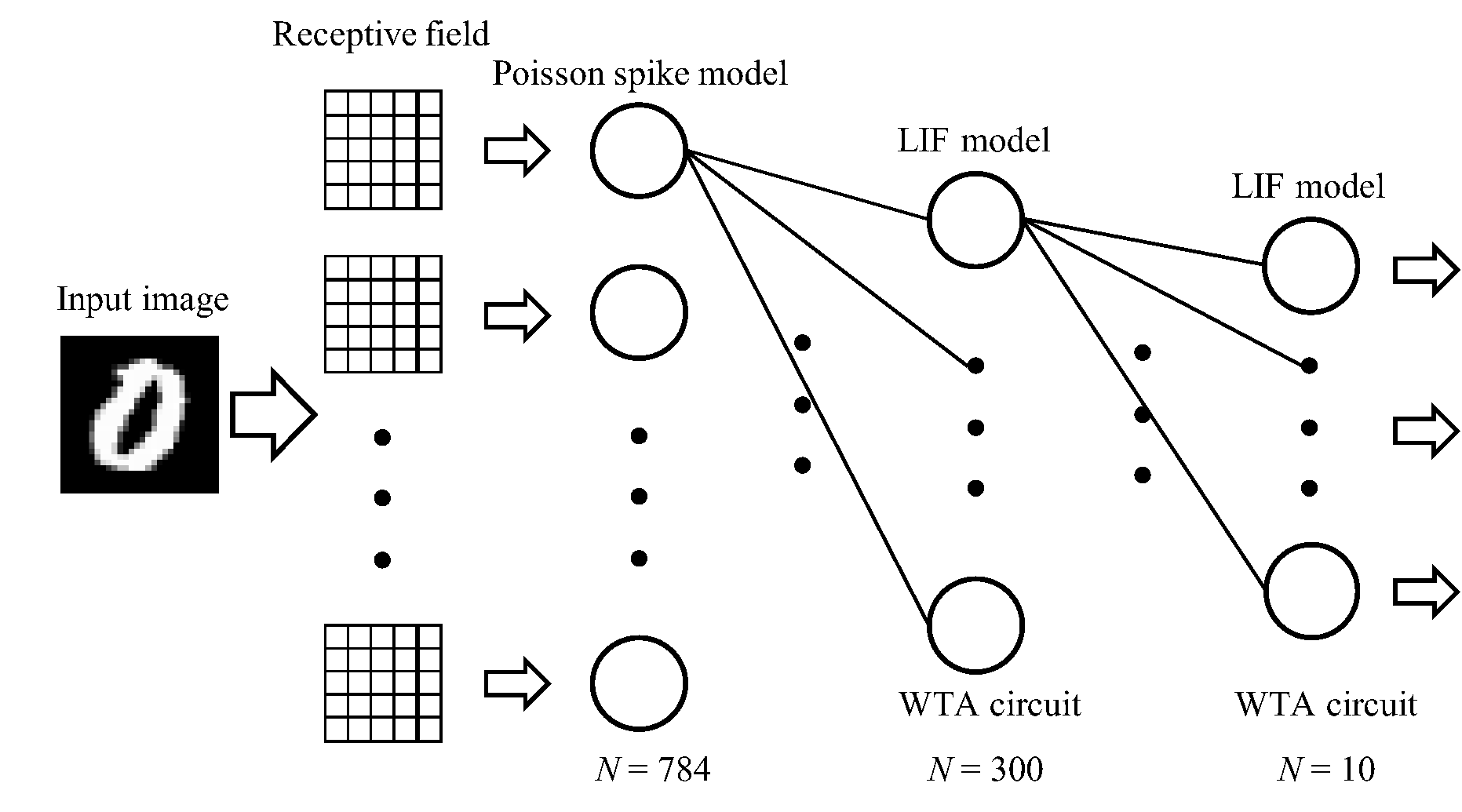}
    \caption{Network architecture.}
    \label{fig:network}
    \end{center}
\end{figure}

\begin{table}[t]
  \begin{center}
  \caption{Values of parameters used in numerical experiments.}
  \begin{tabular}{ll|llll}
  \hline
  \textbf{Used in}         &  & \textbf{Parameter} & \textbf{} & \textbf{Value}    &  \\ \hline
  \multirow{4}{*}{Network} &  & $\tau_{\rm mp}$         &           & 20              &  \\
                           &  & $\mu^{(0)}$         &           & $-0.4$             &  \\
                           &  & $\mu^{(1)}$         &           & $-1.0$              &  \\
                           &  & $\sigma$            &           & 0.5               &  \\ 
                           &  & $T_{ref}$            &           & 1               &  \\ 
                           \hline
  \multirow{5}{*}{BP}      &  & $\alpha$            &           & 2                 &  \\
                           &  & $\eta_{\rm w}$      &           & 0.002             &  \\
                           &  & $\eta_{\rm th}$     &           & $0.1\eta_{\rm w}$ &  \\
                           &  & $\gamma$            &           & 1                 &  \\
                           &  & $\rho$              &           & 0.00004           &  \\ \hline
  \multirow{5}{*}{STDP}    &  & $\tau_{\rm plus}$   &           & 8                 &  \\
                           &  & $\tau_{\rm minus}$  &           & 5                 &  \\
                           &  & $\rm A ^+$          &           & 0.6               &  \\
                           &  & $\rm A^-$           &           & $-0.3$               &  \\
                           &  & $\sigma_{\rm stdp}$ &           & 0.0001            &  \\ \hline
  \label{tab:parameters}
  \end{tabular}
  \end{center}
  \end{table}

\begin{table}[t]
      \begin{center}
      \caption{Computing environment.}
      \begin{tabular}{l|l}
      \hline
          OS       & Ubuntu 18.04.3 LTS         \\ \hline
          CPU      & Intel Core i7-9700K 3.6 GHz \\ \hline
          Memory   & 64 GB                       \\ \hline
          Language & Python 3.7                \\ \hline
          \end{tabular}
          \label{tab:specs}
          \end{center}
\end{table}

\begin{table*}[!t]
  \begin{center}
  \caption{The accuracy and learning time of the proposed method and self-training.}
  \label{tab:proposed and self size10}
  \begin{tabular}{p{2.5cm}|p{2.4cm}|p{2.7cm}|p{2.5cm}|p{1.65cm}}
  \hline
  Method & Best accuracy (\%) & Learning time to best accuracy (s) & Accuracy after overall learning (\%) & Total learning time (s) \\ \hline
  Proposed method          & \textbf{65.2}      & \textbf{2739.334}           & \textbf{60.0}           & \textbf{8500.118}     \\
  Self-training & 63.2       & 18054.187          & 59.5           & 33781.616   \\ \hline
  \end{tabular}
  \end{center}
  \end{table*}

\subsection{Network architecture}
We use a fully connected feed-forward SNN with one hidden layer in the numerical experiments.
The input layer has 784 neurons because we input the image of 28 pixels $\times$ 28 pixels in numerical experiments.
The hidden layer has 300 neurons, and the output layer has 10 neurons in which each neuron corresponds to a correct label from 0 to 9 because we evaluated the result with the classification of 0 to 9 handwritten numeric characters.
The LIF model is used in the hidden and output layers.
In the input layer, the Poisson spike model is used. The input image is convolved with a $5\times5$ on-center receptive field.
In order to match the input size after convolution with the image size before convolution, we perform zero padding, which fills the periphery of the input image with $0$.
The receptive field is weighted to $[-0.5, -0.125, 0.125, 0.625, 1]$ based on the Manhattan distance to the center of the field. 
The firing frequency of each neuron in the input layer is obtained by scaling the normalized pixel value of the convolved image with the maximum firing frequency.
We set the maximum frequency to 150$/\rm s$.
The hidden and output layers have a WTA mechanism.
An outline of the network is shown in Fig. \ref{fig:network}.

\section{Experimental Procedures and Results}
In this section, we present the numerical results obtained by training an SNN using the proposed method.
We evaluated the performance by solving the classification problem of handwritten numeric characters from $0$ to $9$. 
We used the MNIST dataset, which is a grayscale handwritten character image set of $28$ pixels $\times$ $28$ pixels.\cite{MNIST}
In this study, we simulated in discrete time and the timestep was 1 ms.
The input time was 50 ms for learning and 150 ms for testing.
For BP, we used 10 or 30 labeled training data for each number, 100 or 300 in total.
For STDP, we used 500 unlabeled training data for each number, 5000 in total.
For the test data, 10 sets of 10 samples per number, 1000 in total, were used, and the mean of the accuracy for the test datasets was used for evaluation.
The training data for BP, training data for STDP, and the test data were all different.
In the proposed method, learning by BP and STDP is performed as a whole for a total of 200 epochs. 
First, we train the network by BP for 150 epochs, and then learning by STDP is performed for 50 epochs. 
The batch size of BP was 25.
Table \ref{tab:parameters} shows the value of each parameter used in the experiments.
The STDP parameters $\rm A ^+$ and $\rm A ^-$ were set to the same as those in the previous research.\cite{iakymchuk2015simplified}
Although a grid search was performed, we finally found that these parameters in the previous research are enough to stabilize the learning process.
The learning rate of STDP was set small enough for learning to be stable in the example here.
The refractory period is not considered because we set $T_{\rm ref}=1$ ms and the timestep is also 1 ms. 
This approximation is reasonable because the activation of neurons in SNNs is rarely dominated by the refractory period.\cite{Lee_SpikingNNBackprop}
We evaluate the test data for each epoch and compare the result of learning by only BP with that of learning by the proposed method. 
In addition, the results of learning by self-training, which is the existing semi-supervised method, are also compared with those of learning by the proposed method. 
The mean for 10 sets of test data in each epoch was used as the accuracy, and a standard deviation of $1\sigma$ was used for confidence intervals. 
The computing environment used in experiments is shown in Table \ref{tab:specs}.

\subsection{Less labeled data: 10 $\times$ 10 samples}
\begin{figure}[tb]
    \begin{center}
    \includegraphics[scale=0.39]{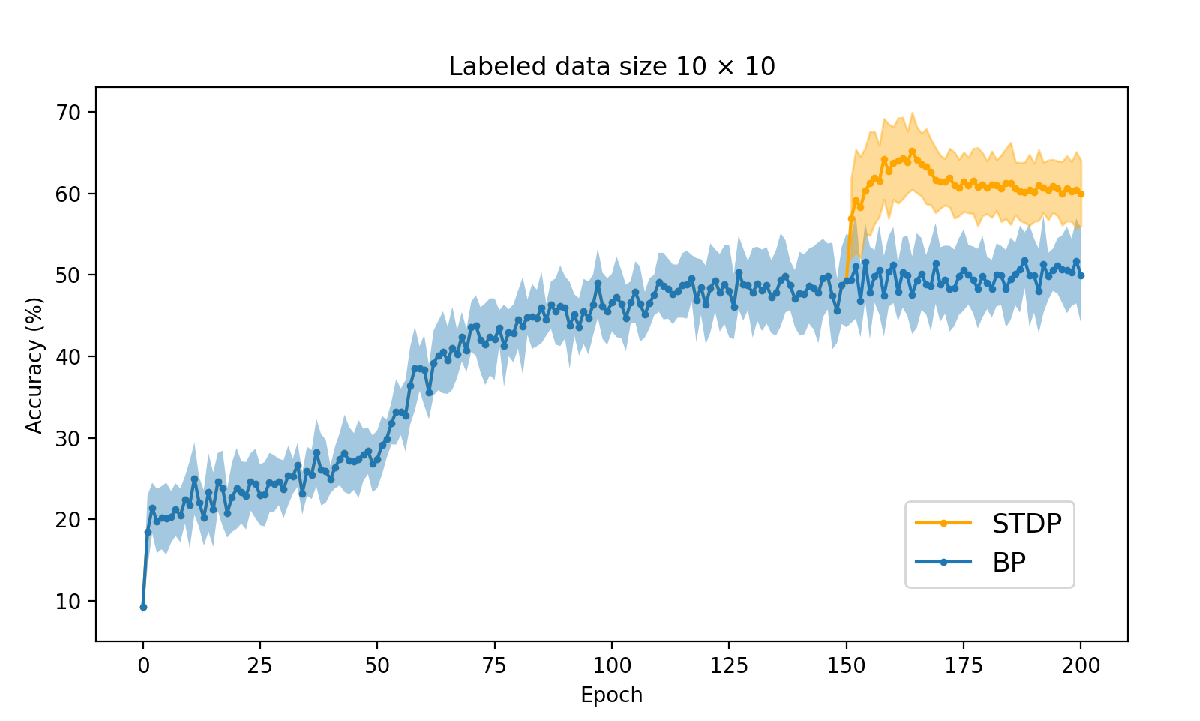}
    \caption{(Color online) Plot of the accuracy against epoch ($10\times10$ samples labeled data).}
    \label{fig:BPsize10}
    \end{center}
\end{figure}

\begin{figure}[tb]
    \begin{center}
    \includegraphics[scale=0.28]{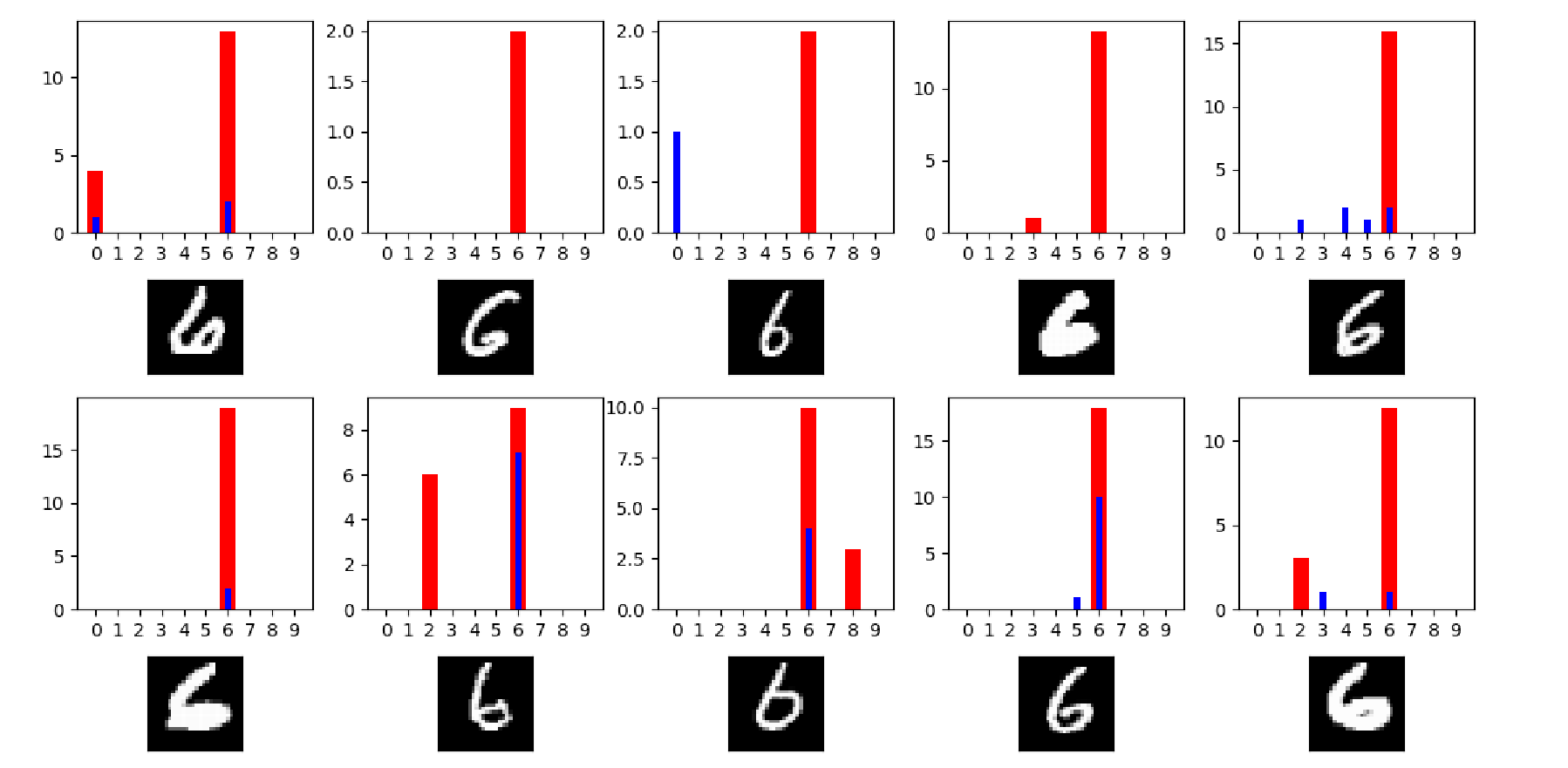}
    \caption{(Color online) Number of firings before and after STDP learning. The blue bar denotes the number of firings before STDP learning and the red bar denotes the number of firings after STDP learning.}

    \label{fig:fir_num}
    \end{center}
\end{figure}

Figure \ref{fig:BPsize10} shows the results when $10$ training samples $\times\ 10$ digits are used as the labeled training data for BP. 
This demonstrates that the accuracy is improved by STDP-based unsupervised learning after BP-based supervised learning, compared with the case of learning by only BP.
Figure \ref{fig:fir_num} shows the number of firings of the output layer for the test data when using only BP and when using the proposed method. 
STDP modifies the network adequately; the number of correct firings increases even when the BP-based supervised learning yields incorrect results.

\begin{figure}[t]
    \begin{center}
    \includegraphics[scale=0.39]{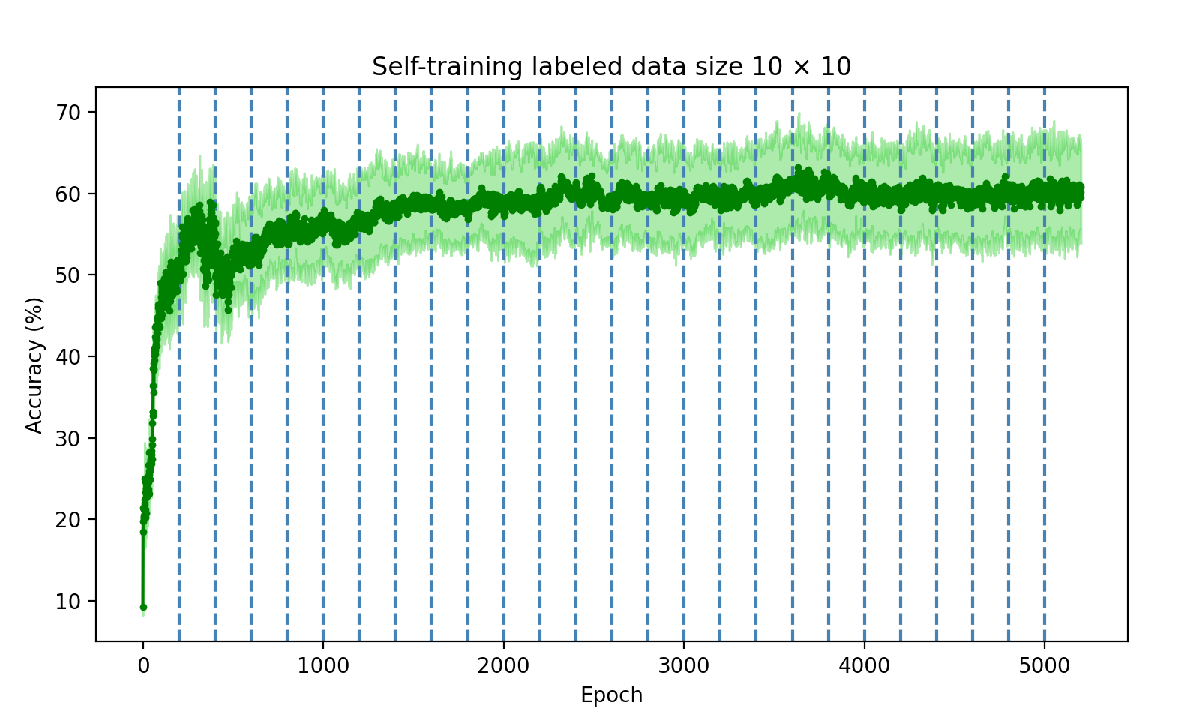}
    \caption{(Color online) Plot of the accuracy against epoch of self-training ($10\times10$ samples labeled data). The blue dotted lines show the labeling steps.}
    \small{}
    \label{fig:self_training_size10}
    \end{center}
\end{figure}

Next, we compared the results of the proposed method with those of learning by self-training. 
In self-training, two processes, learning for 200 epochs by BP and labeling of unlabeled data, were repeated until all unlabeled data were labeled. 
In the labeling step, unlabeled data was input for 150 ms, and then we labeled the top 200 data in which neurons in the output layer fired most frequently among the unlabeled data. 
Figure \ref{fig:self_training_size10} shows the result of learning by self-training. 
We can see that the accuracy does not improve much even if the amount of labeled data increases because of mislabeling.
Table \ref{tab:proposed and self size10} shows the accuracy and the learning time for the proposed method and self-training. The accuracy of the proposed method is higher than that of self-training, and the learning time of the proposed method is shorter than that of self-training.

\begin{figure}[t]
  \begin{center}
  \includegraphics[scale=0.39]{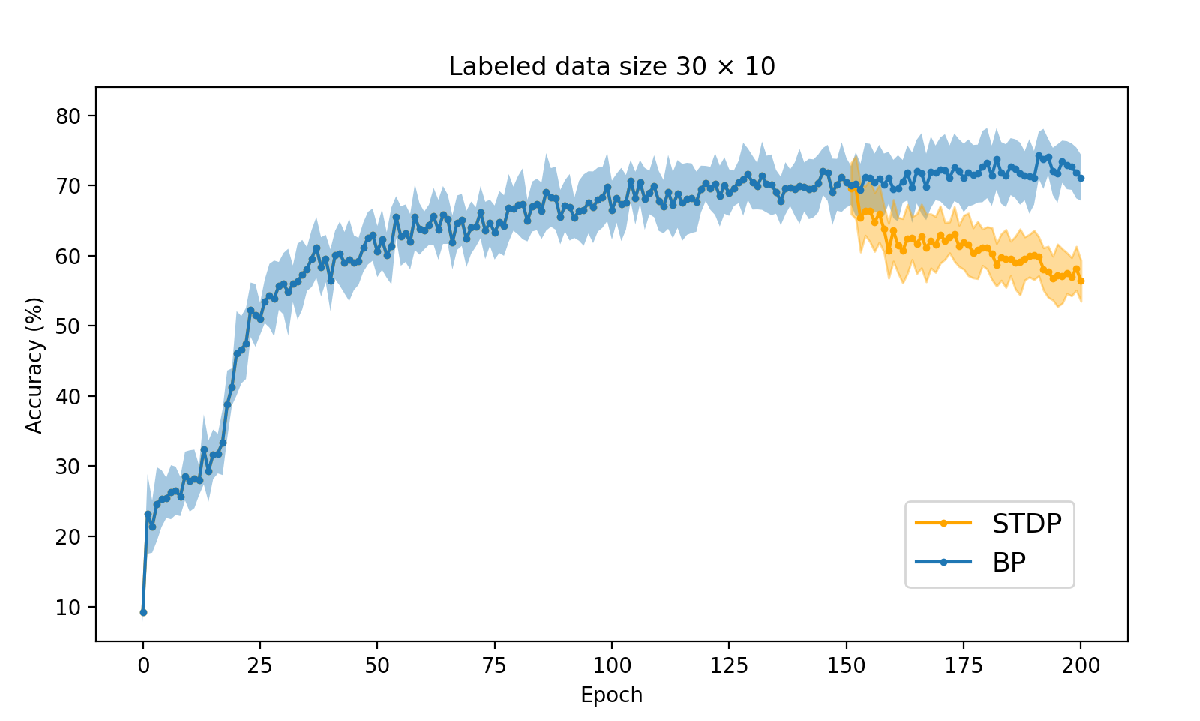}
  \caption{(Color online) Plot of the accuracy against epoch ($30\times10$ samples labeled data).}
  \small{}
  \label{fig:BPsize30}
  \end{center}
\end{figure}

\begin{figure}[t]
  \begin{center}
  \includegraphics[scale=0.39]{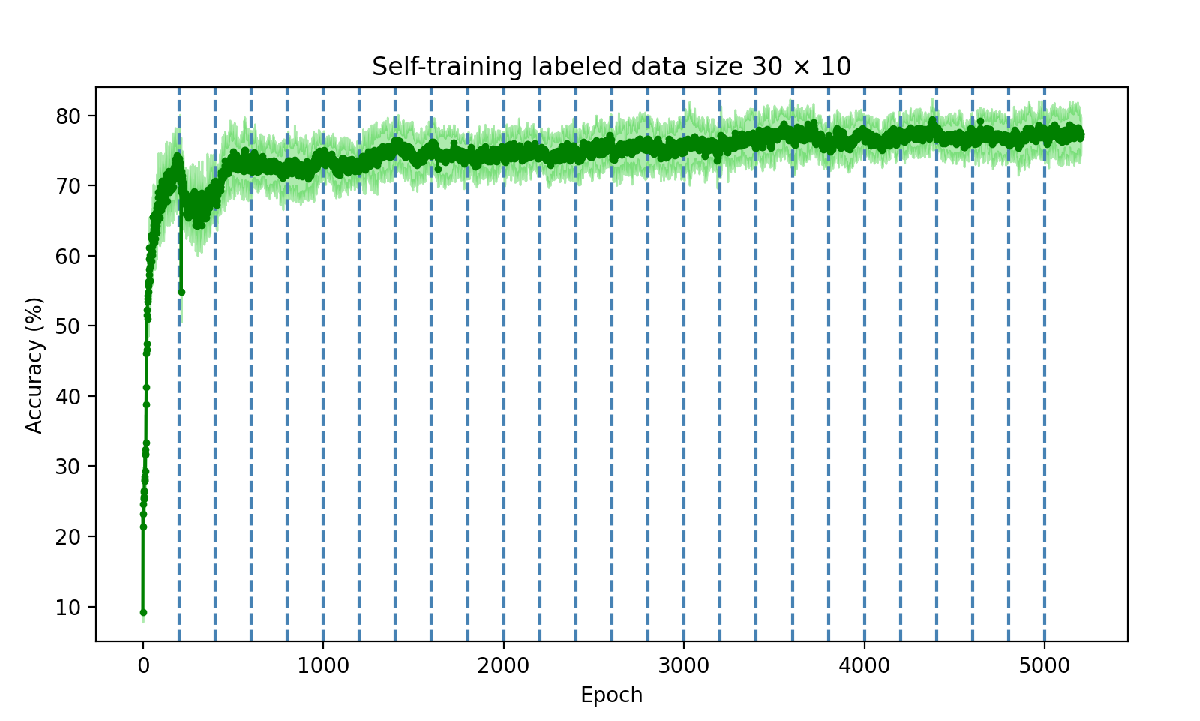}
  \caption{(Color online) Plot of the accuracy against epoch of self-training ($30\times10$ samples labeled data). The blue dotted lines show the labeling steps.}
  \small{}
  \label{fig:self_training_size30}
  \end{center}
\end{figure}

\subsection{More labeled data: 30 $\times$ 10 samples}
Figure \ref{fig:BPsize30} shows the results when $30$ training samples $\times\ 10$ digits are used as labeled training data for BP.
When the amount of labeled training data is as large as $30\times10$ samples, it can be seen that the combination of BP and STDP decreases the accuracy.

Next, we compared the results with those of learning by self-training. 
Figure \ref{fig:self_training_size30} shows the result of learning by self-training. 
Because there is a lot of labeled training data, mislabeling does not occur at such a high rate and the accuracy is improved.

\section{Discussion}\begin{figure}[tb]
  \begin{center}
  \includegraphics[scale=0.4]{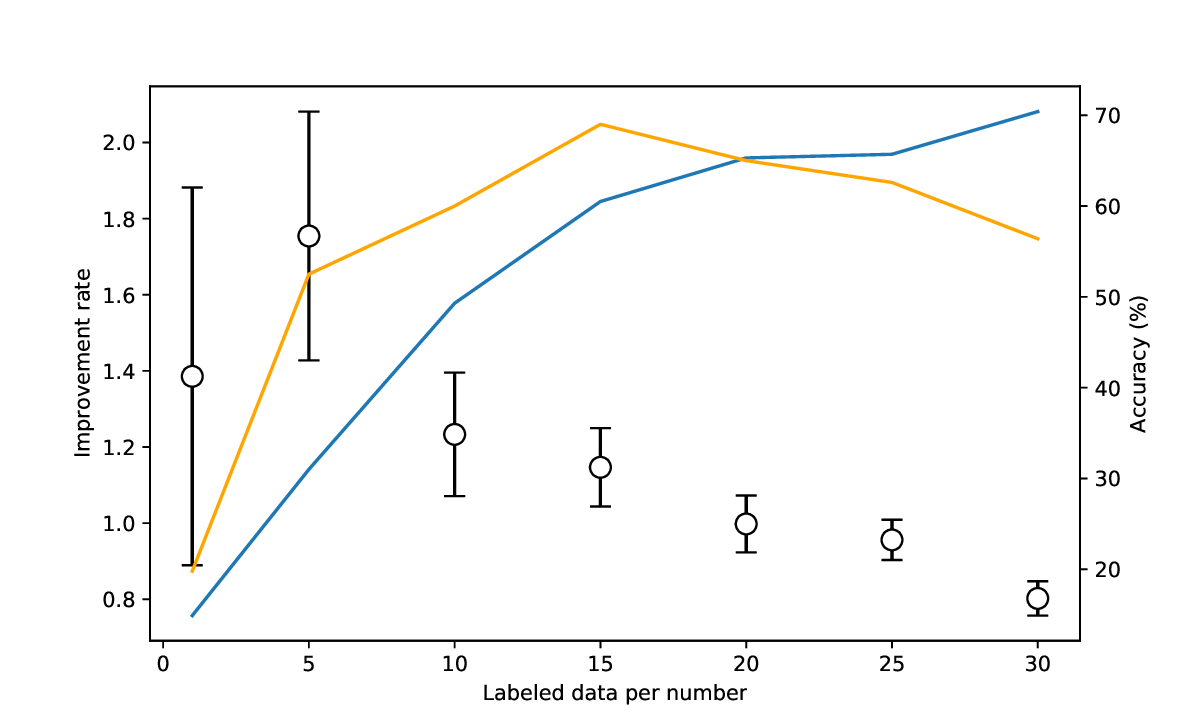}
  \caption{(Color online) Plot of the improvement rate by applying STDP and the accuracy after applying BP and STDP against the amount of labeled data. Each white circle indicates the mean improvement rate of 10 sets of test data, and a standard deviation of $1\sigma$ is used for the error bar. The improvement rate is defined as the ratio of the accuracy after applying STDP to that before it. The blue and orange lines correspond to the accuracy at the end of the BP stage and that after the STDP learning stage, respectively.}
  \small{}
  \label{fig:improvement rate}
  \end{center}
\end{figure}

The results of simulations using our proposed semi-supervised learning method demonstrate that using it can improve the accuracy of SNNs when there is a small amount of labeled data.
Moreover, in this study, we did not need to label unlabeled data.
Thus, the problem that learning is not performed well due to mislabeling in the existing semi-supervised learning method for discriminative models is solved. This method can be applied effectively when there are few labeled training data points and the accuracy is low.
In addition, the learning time can be reduced because it does not need to re-learn like self-training and consists of only two steps: BP learning and STDP learning.
Because the number of epochs to reach the best accuracy is lower than that in self-training, unlabeled training data can be used for efficient learning.
On the other hand, when there is a relatively large amount of labeled data, the accuracy decreases.
Hence, this method has the property that when the amount of labeled training data is extremely small, the performance can be improved, but when the amount of labeled training data is relatively large, the accuracy can decrease.
Such a feature has been reported in an existing semi-supervised learning method as well,\cite{Tagging_English_Text_semi-supervised} and this supports the fact that semi-supervised learning is being performed in this work.

\begin{figure*}[tb]
  \begin{tabular}{cc}
    \begin{minipage}[t]{0.45\hsize}
      \centering
      \includegraphics[keepaspectratio, scale=0.4]{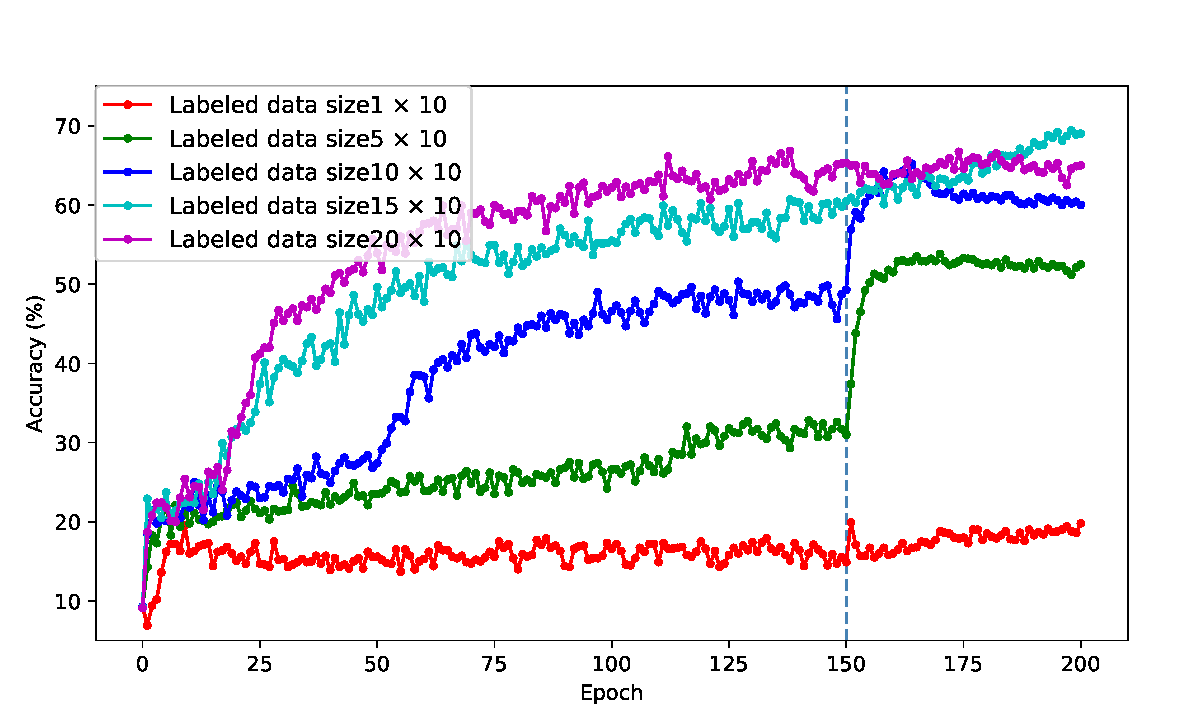}
      \caption{(Color online) Plot of the accuracy for the cases in which the mean improvement rates are greater than or equal to 1. The blue dotted line indicates the epoch to start learning by STDP.}
      \label{fig:is1 or greater}
    \end{minipage} &
    \hspace{7mm}
    \begin{minipage}[t]{0.45\hsize}
      \centering
      \includegraphics[keepaspectratio, scale=0.4]{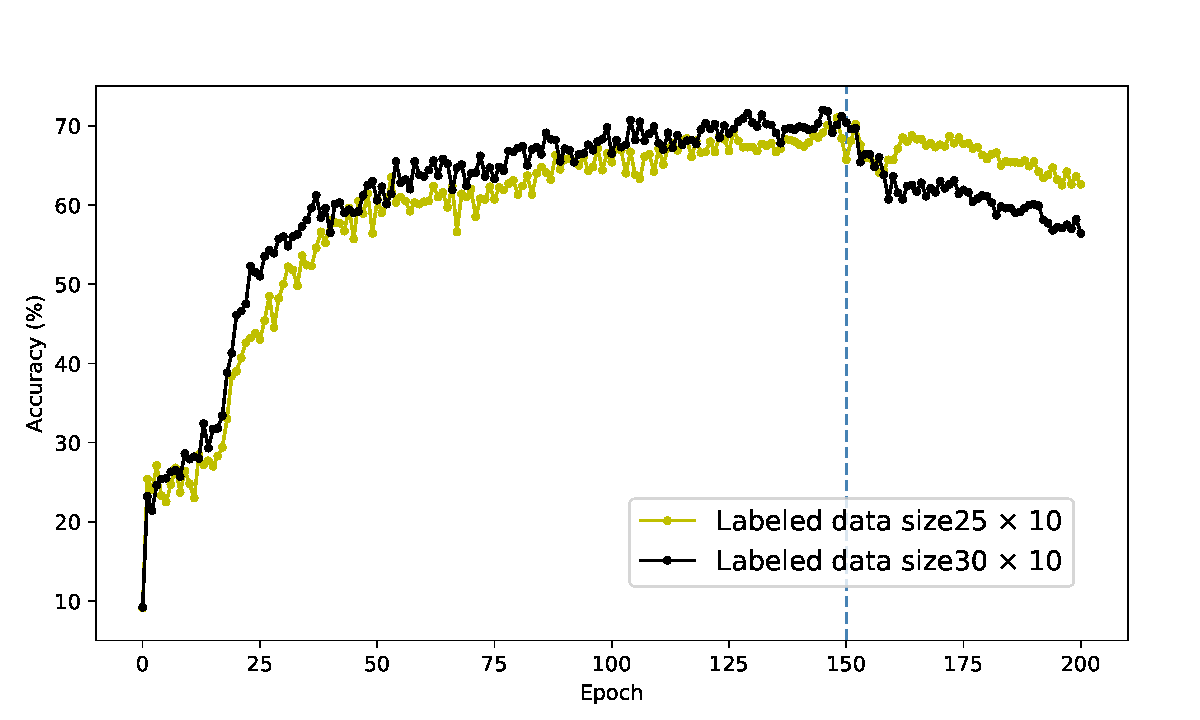}
      \caption{(Color online) Plot of the accuracy for the cases in which the mean improvement rates are less than 1. The blue dotted line indicates the epoch to start learning by STDP.}
      \label{fig:less than 1}
    \end{minipage}
  \end{tabular}
\end{figure*}

Pre-training by unsupervised learning has been interpreted as a type of self-organization,\cite{ohzeki2015statistical} and STDP is considered to be involved in self-organization in living organisms.\cite{effenberger2015Self_oragnizetion_by_STDP}
Therefore, the method of pre-training by STDP followed by fine-tuning with BP \cite{SNN_STDP_followedby_BPFineTunig,Semi-Supervised_BP_after_STDP} can be interpreted as the sequence of learning the structure of input data by self-organization and then performing supervised learning by BP.
In contrast, from our results, it can be inferred that STDP plays an important role in learning other than self-organization, such as fine-tuning or confirmation, and thus the result is interesting from an engineering perspective as well as a biological one.

Although we have discussed the proposed method from a biological point of view, there are various arguments about the biological plausibility of BP.\cite{towardsBP}
In this research, to focus on the effect of the application of STDP after BP, we used the BP method of Lee et al.,\cite{Lee_SpikingNNBackprop} which has high performance in SNNs but does not seem to be biologically plausible.
Using the supervised STDP\cite{supervisedSTDP,BP-STDP} would allow us to devise a more biologically plausible learning scheme.
In addition, we can develop an end-to-end STDP-based learning method for SNNs by combining STDP-based pre-training\cite{SNN_STDP_followedby_BPFineTunig,Semi-Supervised_BP_after_STDP}  with these methods.

From an engineering point of view, it is important to judge whether the proposed method should be applied or not because it can induce the improvement but the deterioration in some cases.
Figure \ref{fig:improvement rate} shows the improvement rate before and after applying STDP against the amount of labeled data.
Figure \ref{fig:is1 or greater} shows the change in accuracy for the cases in which the improvement rates are greater than or equal to 1, and Fig. \ref{fig:less than 1} shows the change in accuracy for the cases in which the improvement rates are less than 1.
From these results, the degradation of accuracy basically occurs only when there is a relatively large amount of labeled data.
Note that the degradation can also occur when labeled data per number is extremely small because the standard deviation of improvement rate is quite large and thus the improvement rate can take values less than 1.
This is because the accuracy after learning by BP is quite low.
When the amount of training data is extremely small like 1 $\times$ 10 samples, the learning by BP can occasionally fail because it is highly dependent on the training data.
In this case, we get into the situation where the learning is performed only by STDP, which is unsupervised learning, and the accuracy will decrease with the evaluation criterion of supervised learning.
Overall, it is easy to detect the degradation after the application by STDP.
Therefore, for engineering purposes, we should monitor the change in accuracy for a few epochs after applying STDP, and if a deterioration is observed, we can terminate the training like early stopping.\cite{Earlystopping}

Although we simulated in discrete time in this study, all the algorithms and architectures used in our work can be event-driven. Thus, the proposed method can be applied in an event-driven manner. 
We expect that it would be highly efficient in real-world problems if implemented in neuromorphic hardware.

\section{Conclusions}
We proposed a semi-supervised learning method with unsupervised learning by STDP after supervised learning by BP.
This method reverses the order of the learning steps in a previously reported semi-supervised learning method.\cite{SNN_STDP_followedby_BPFineTunig,Semi-Supervised_BP_after_STDP} We developed the proposed method based on its parallel to the biological situation of an organism learning a label and subsequently observing its environment to confirm this learning.
The results of numerical simulations show that our proposed method displays good accuracy, particularly when only a small amount of labeled data is used. However, when a relatively large amount of labeled data is used, the accuracy decreases.
Our results also show that STDP plays an important role in learning other than self-organization, such as fine-tuning or confirmation. 

This research was devised from an intuitive sense of organisms and the learning rules that have been discovered in them, and mathematical verification is required to understand why STDP improves accuracy. 
Moreover, now there are three types of STDP-based learning methods; STDP-based pre-training,\cite{SNN_STDP_followedby_BPFineTunig,Semi-Supervised_BP_after_STDP} supervised STDP learning,\cite{supervisedSTDP,BP-STDP} and our method. Thus, by combining these methods, we can develop an end-to-end STDP-based learning method for SNNs.
Such verification and expansion can lead to an improved understanding of the brain and efficient processing in real-time.


\begin{thebibliography}{10}
\bibitem{NeuronalDynamics}
W.~Gerstner, W.~M. Kistler, R.~Naud, and L.~Paninski, {\em Neuronal Dynamics:
  From Single Neurons to Networks and Models of Cognition} (Cambridge
  University Press, Cambridge, 2014).

\bibitem{lecun2015deep}
Y.~LeCun, Y.~Bengio, and G.~Hinton, Nature {\bfseries 521},  436 (2015).

\bibitem{Efficient_BP_Neuromophic}
S.~K. Esser, R.~Appuswamy, P.~A. Merolla, J.~V. Arthur, and D.~S. Modha,
  NIPS'15: Proc. 28th Int. Conf. Neural Information Processing
  Systems, 2015, Vol. 1, Montreal, p. 1117.

\bibitem{maass1997SNN}
W.~Maass, Neural Netw. {\bfseries 10},  1659 (1997).

\bibitem{Review_deepL_inSNN}
A.~Tavanaei, M.~Ghodrati, S.~R. Kheradpisheh, T.~Masquelier, and A.~Maida, Neural Netw. {\bfseries 111},  47 (2019).

\bibitem{Guest_2018}
D.~Guest, K.~Cranmer, and D.~Whiteson, Annu. Rev. Nucl. Part. Sci. {\bfseries
  68},  161 (2018).

\bibitem{Borzyszkowski:2687102}
B.~P. Borzyszkowski,  presented at {Neuromorphic Computing in High Energy
  Physics. Second CERN openlab summer student lightning talk session},  2019.

\bibitem{maass2004computational}
W.~Maass and H.~Markram, J. Comput. Syst. Sci. {\bfseries 69},  593 (2004).

\bibitem{SpikeProp}
S.~M. Bohte, J.~N. Kok, and H.~La~Poutre, Neurocomputing {\bfseries 48},  17
  (2002).

\bibitem{TrainingDeepSNN2020}
E.~Ledinauskas, J.~Ruseckas, A.~Jur{\v{s}}{\.e}nas, and G.~Bura{\v{c}}as,
  arXiv:2006.04436.

\bibitem{TemporalBackprop}
H.~Mostafa, IEEE Trans. Neural Netw. Learn. Syst. {\bfseries 29},  3227 (2017).

\bibitem{Lee_SpikingNNBackprop}
J.~H. Lee, T.~Delbruck, and M.~Pfeiffer, Front. Neurosci. {\bfseries 10},  508
  (2016).

\bibitem{STDP_Overview}
H.~Markram, W.~Gerstner, and P.~J. Sj{\"o}str{\"o}m, Front. Synaptic Neurosci.
  {\bfseries 4},  2 (2012).

\bibitem{iakymchuk2015simplified}
T.~Iakymchuk, A.~Rosado-Mu{\~n}oz, J.~F. Guerrero-Mart{\'\i}nez,
  M.~Bataller-Mompe{\'a}n, and J.~V. Franc{\'e}s-V{\'\i}llora, EURASIP J. Image
  Video Process. {\bfseries 2015},  4 (2015).

\bibitem{STDP_MNIST}
P.~U. Diehl and M.~Cook, Front. Comput. Neurosci. {\bfseries 9},  99 (2015).

\bibitem{supervisedSTDP}
Y.~{Zeng}, K.~{Devincentis}, Y.~{Xiao}, Z.~I. {Ferdous}, X.~{Guo}, Z.~{Yan},
  and Y.~{Berdichevsky}, 2018 IEEE Int. Conf. Acoustics, Speech and Signal
  Processing, 2018, Alberta, p. 1154.

\bibitem{BP-STDP}
A.~Tavanaei and A.~Maida, Neurocomputing {\bfseries 330},  39  (2019).

\bibitem{semi-supervisedReview}
Y.~P. Reddy, P.~Viswanath, and B.~E. Reddy, Int. J. Eng. Technol. Innov.
  {\bfseries 7},  81 (2018).

\bibitem{survey_suemi-supervised}
J.~E. Van~Engelen and H.~H. Hoos, Machine Learning {\bfseries 109},  373
  (2020).

\bibitem{erhan2010does}
D.~Erhan, A.~Courville, Y.~Bengio, and P.~Vincent, Proc. Thirteenth Int.
  Conf. Artificial Intelligence and Statistics, 2010, Vol. 9, Sardinia, p.
  201.

\bibitem{hinton2006fast}
G.~E. Hinton, S.~Osindero, and Y.-W. Teh, Neural Comput. {\bfseries 18},  1527
  (2006).

\bibitem{SNN_STDP_followedby_BPFineTunig}
C.~Lee, P.~Panda, G.~Srinivasan, and K.~Roy, Front. Neurosci. {\bfseries 12},
  435 (2018).

\bibitem{Semi-Supervised_BP_after_STDP}
Y.~{Dorogyy} and V.~{Kolisnichenko}, 2018 IEEE First Int. Conf. System
  Analysis Intelligent Computing, 2018, Kiev, p.~1.

\bibitem{oster2009WinnerTakeALL}
M.~Oster, R.~Douglas, and S.-C. Liu, Neural Comput. {\bfseries 21},  2437
  (2009).

\bibitem{gerstner2002SpikingNeuronModel}
W.~Gerstner and W.~M. Kistler, {\em Spiking Neuron Models: Single Neurons,
  Populations, Plasticity} (Cambridge University Press, Cambridge, 2002).

\bibitem{riesenhuber1999hierarchical}
M.~Riesenhuber and T.~Poggio, Nat. Neurosci. {\bfseries 2},  1019 (1999).

\bibitem{STDP_formulation}
S.~Song, K.~D. Miller, and L.~F. Abbott, Nat. Neurosci. {\bfseries 3},  919
  (2000).

\bibitem{MNIST}
Y.~LeCun, L.~Bottou, Y.~Bengio, and P.~Haffner, Proc. IEEE {\bfseries 86},
  2278 (1998).

\bibitem{STDP_TimeWindow}
L.~I. Zhang, H.~W. Tao, C.~E. Holt, W.~A. Harris, M.~Poo, Nat. {\bfseries 395}, 37 (1998).

\bibitem{Tagging_English_Text_semi-supervised}
B.~Merialdo, Comput. Linguist. {\bfseries 20},  155 (1994).

\bibitem{ohzeki2015statistical}
M.~Ohzeki, J. Phys. Soc. Jpn. {\bfseries 84},  034003 (2015).

\bibitem{effenberger2015Self_oragnizetion_by_STDP}
F.~Effenberger, J.~Jost, and A.~Levina, PLoS Comput. Biol. {\bfseries 11},
  (2015).

\bibitem{towardsBP}
Y.~Bengio, D.~H. Lee, J.~Bornschein, T.~Mesnard and Z.~Lin, 
arXiv:1502.04156.

\bibitem{Earlystopping}
L.~Prechelt, Neural Netw.:Tricks of the trade. {\bfseries 1524}, 55 (1998).

\end{thebibliography}
\end{document}